\documentclass[letterpaper, 10 pt, conference]{ieeeconf}  
\usepackage{graphicx}
\usepackage{mathtools}
\usepackage{cleveref}
\usepackage{amsmath}

\newcommand{\dl}{\Delta l}

\IEEEoverridecommandlockouts                              

\title{\LARGE \bf
Addition of a peristaltic wave improves multi-legged locomotion performance on complex terrains}

\author{Massimiliano Iaschi$^{1,*}$, Baxi Chong$^{1,*}$, Tianyu Wang$^{1}$, Jianfeng Lin$^{1}$, Juntao He$^{1}$ \\  Daniel Soto$^{1}$, Zhaochen Xu$^{1}$, Daniel I Goldman$^{1}$
\thanks{*These authors contributed equally to this work}
\thanks{$^{1}$Massimiliano Iaschi, Baxi Chong, Tianyu Wang, Jianfeng Lin, Daniel Soto, and Daniel I Goldman are with the Georgia Institute of Technology, Atlanta, GA 30332.
        {\tt\small \{miaschi3, bchong9, tianyuwang, jianf.lin, dsoto7\}@gatech.edu, {\tt\small daniel.goldman@physics.gatech.edu}}}%
}

\begin{document}

\maketitle
\thispagestyle{empty}
\pagestyle{empty}

\begin{abstract}
Characterized by their elongate bodies and relatively simple legs, multi-legged robots have the potential to locomote through complex terrains for applications such as search-and-rescue and terrain inspection. Prior work has developed effective and reliable locomotion strategies for multi-legged robots by propagating the two waves of lateral body undulation and leg stepping, which we will refer to as the two-wave template. However, these robots have limited capability to climb over obstacles with sizes comparable to their heights. We hypothesize that such limitations stem from the two-wave template that we used to prescribe the multi-legged locomotion. Seeking effective alternative waves for obstacle-climbing, we designed a five-segment robot with static (non-actuated) legs, where each cable-driven joint has a rotational degree-of-freedom (DoF) in the sagittal plane (vertical wave) and a linear DoF (peristaltic wave). We tested robot locomotion performance on a flat terrain and a rugose terrain. While the benefit of peristalsis on flat-ground locomotion is marginal, the inclusion of a peristaltic wave substantially improves the locomotion performance in rugose terrains: it not only enables obstacle-climbing capabilities with obstacles having a similar height as the robot, but it also significantly improves the traversing capabilities of the robot in such terrains. Our results demonstrate an alternative actuation mechanism for multi-legged robots, paving the way towards all-terrain multi-legged robots.

\end{abstract}

\section{INTRODUCTION}

\begin{figure}[t]
    \centering
    \includegraphics[width=\linewidth]{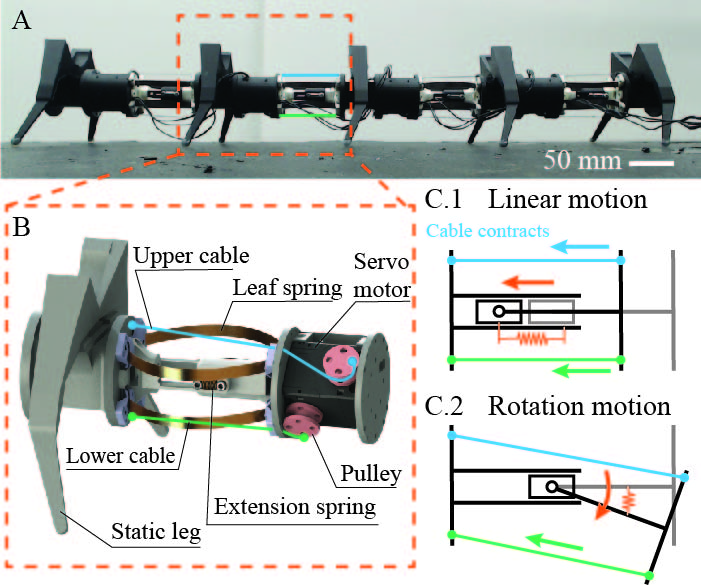}
    \caption{\textbf{Overview of multi-segment robot that locomotes using a combination of peristaltic and vertical waves.} (A) A side-view picture of the robot in resting position on the flat terrain. (B) Computer Aided Design (CAD) zoomed-in representation of a robot two-DoF joint. (C) Side-view diagram showing the actuation mechanism for each DoF.}
    \label{fig:fig1}
\end{figure}%

Multi-legged robots, characterized by multi-segment bodies and relatively simple legs, offer an alternative approach to effective and reliable terrestrial locomotion, contrasting with the conventional bipedal and quadrupedal robots~\cite{chongScience,ozkansystematic,chong2022general,chong2023self}. Specifically, the redundancy in morphology (the many legs) contributes to locomotion robustness: with sufficient legs, it is possible to locomote on complex terrains without additional sensing and control~\cite{chongScience}. Thus, this design paradigm shifts the complexity of legged locomotion from real-time detection and response to the coordination of the multiple degrees-of-freedom in the many modules.

Previous studies have successfully demonstrated multi-legged locomotion strategies that employ propagating waves of lateral body undulation and leg stepping from head to tail~\cite{chong2023self}. However, these robots face challenges when encountering tall obstacles with heights comparable to their own size~\cite{he2024control}. In particular, the challenges are two-fold: (1) a lack of obstacle-climbing capabilities, and (2) insufficient thrust generation off the interaction with tall obstacles for self-propulsion.

We propose that these limitations stem from the specific waves we used to coordinate multi-legged locomotion. Notably, diverse forms of body waves observed in natural organisms suggest potential alternative locomotion strategies. For example, caterpillars exhibit remarkable terrain-negotiating capabilities~\cite{Trimmer2006CaterpillarLA} through a combination of periodic body contraction and vertical body undulation~\cite{cater}. In snakes, vertical waves have been shown to contribute significantly to thrust generation in obstacle-rich environments~\cite{fu2022snakes}. 

To investigate effective alternative waves for obstacle navigation, we designed a five-segment cable-driven multi-legged robot, in which each segment features two degrees-of-freedom (DoF) that allow the robot to move by means of a combination of a vertical and a peristaltic wave (Fig.~1). We show that appropriate phasing of vertical and peristaltic waves facilitates effective forward movement, achieving $0.25 \pm 0.02$ body lengths per cycle (BL/cyc) on flat and hard ground, compared to $0.15 \pm 0.02$ BL/cyc with inappropriate phasing. When tested on rugose terrains, the inclusion of peristaltic waves not only enabled the robot to climb obstacles comparable to its own height, but also significantly enhanced the overall locomotion performance on complex terrains by adjusting the head trajectory and therefore avoiding detrimental interactions with obstacles (``jamming" between obstacles).

\begin{figure}
    \centering
    \includegraphics[width=\linewidth]{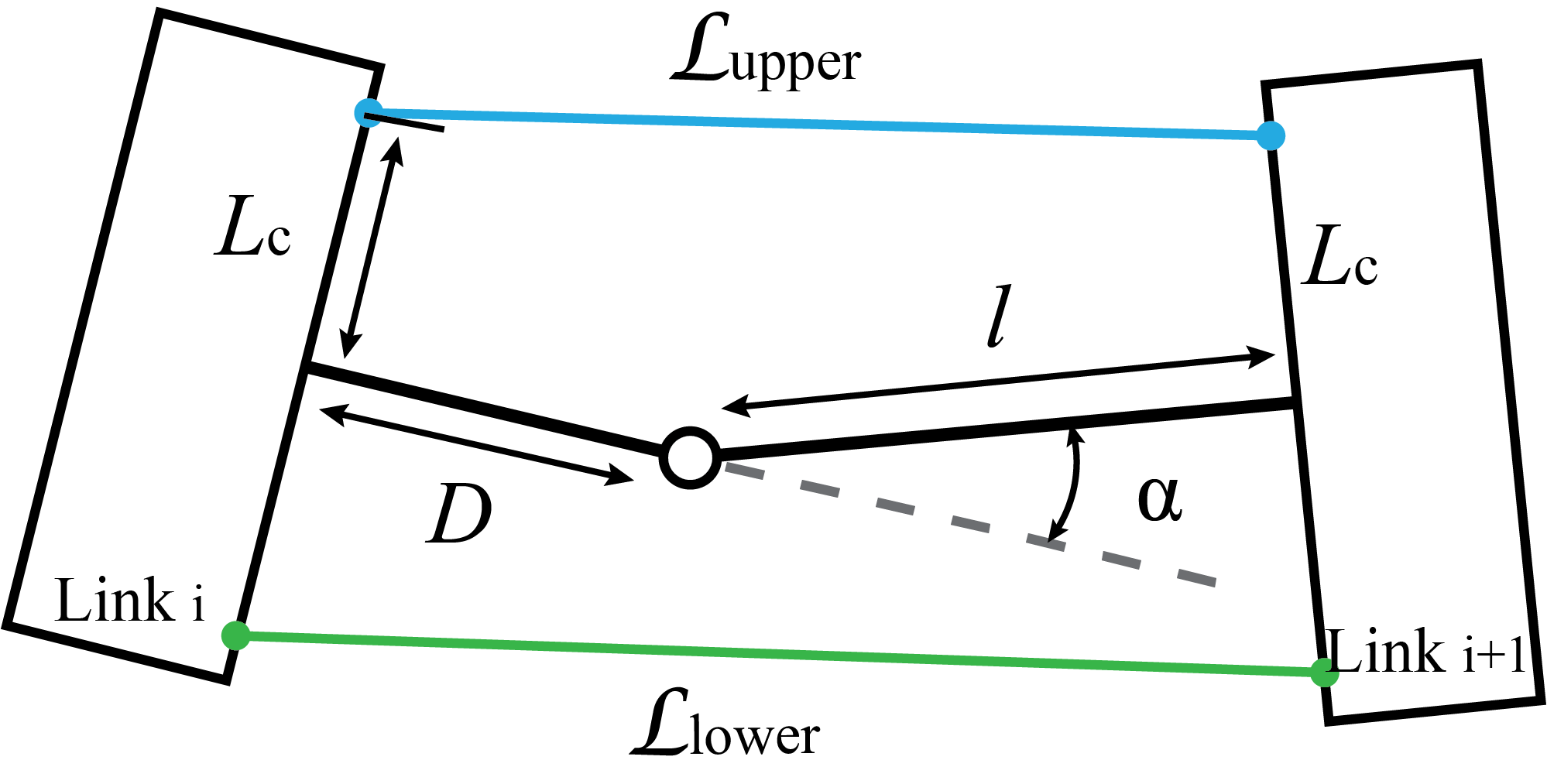}
    \caption{\textbf{Geometry of an individual joint.}  $\mathcal{L}_{upper}$ and $\mathcal{L}_{lower}$ refer to the length of the upper and lower cables. $D$ and $L_c$ are dimensional constants of the robot. $l$ and $\alpha$ are the peristaltic length and the joint angle controlled by the gait equation, respectively.}
        \label{fig:geometric}
\end{figure}%

\section{ROBOT DESIGN AND CONTROL}

\subsection{Module Components}

Most robots with body contraction/extension capabilities employ techniques from soft robotics, such as silicone bodies with pneumatic actuation \cite{mazzolai,8722821}, flexible braided mesh tubes with NiTi coil actuators \cite{5509542}, or braided meshes with cable-driven actuation mechanisms \cite{doi:10.1177/0278364911432486}. However, despite all the benefits of compliance dynamics, these soft parts introduce substantial challenges to precise joint control and actuation. Specifically, the soft-actuation mechanisms suffer from (1) insufficient torque generation, (2) delay from torque-generation to joint-actuation, and (3) history-dependence on the joint-actuation~\cite{lee2017soft}.  

Recently, hybrid designs consisting of rigid bodies coupled with soft actuation (e.g., cable-driven), compliant elements (e.g., springs or spring steel elements), or both, were proposed to simplify snake-inspired and earthworm-inspired robot locomotion~\cite{wang2023mechanical,kojouharov2024anisotropic,Omori2009AnUE,BI2023108436}. Such hybrid design imposes time-dependent constraints on the joint angle, which enables the active transition between rigid-actuation and soft-actuation. 

In this paper, we propose an hybrid cord-driven robot (Fig.~1A) consisting of a series of 5 modules (75 cm total length) connected by 4 joints capable of both vertical and peristaltic motion, combined with compliant components (Fig.~1B). Each joint presents two degrees of freedom (DoFs): a rotational DoF in the sagittal plane and a linear DoF for periodic compression and extension. The rotational DoF, actuated by introducing a differential in the upper and lower cord lengths as shown in Fig.~1C.2, composes the vertical wave, while the linear DoF, actuated by simultaneously compressing (Fig.~3A.2) or relaxing (Fig.~3A.1) the upper and lower cords as shown in Fig. 1C.1, composes the peristaltic wave. 

To facilitate a comparative analysis on the role of the peristaltic wave, we designed a pitch-only joint (Fig.~3C.2) to contrast the two-DoF joint (Fig.~3.C.1). The pitch-only joint is made up of two components (3D-printed in PLA), with a metal pin that passes through their aligned holes allowing for their planar rotation. The main difference between the pitch-only joint and the two-DoF joint lies in the hinge connection. The two-DoF joint hinge connection has a slot that enables the active compression, whereas the pitch-only joint has a rigid connection with a fixed joint length (therefore no compression). In both cases the dimensions of the module are identical (6.5cm in diameter and 8.75cm in length). Throughout this paper we will refer to the two joints as the two-DoF joint (Fig.~3C.1) and the pitch-only joint (Fig.~ 3C.2).

 \begin{figure}
    \centering
    \includegraphics[width=\linewidth]{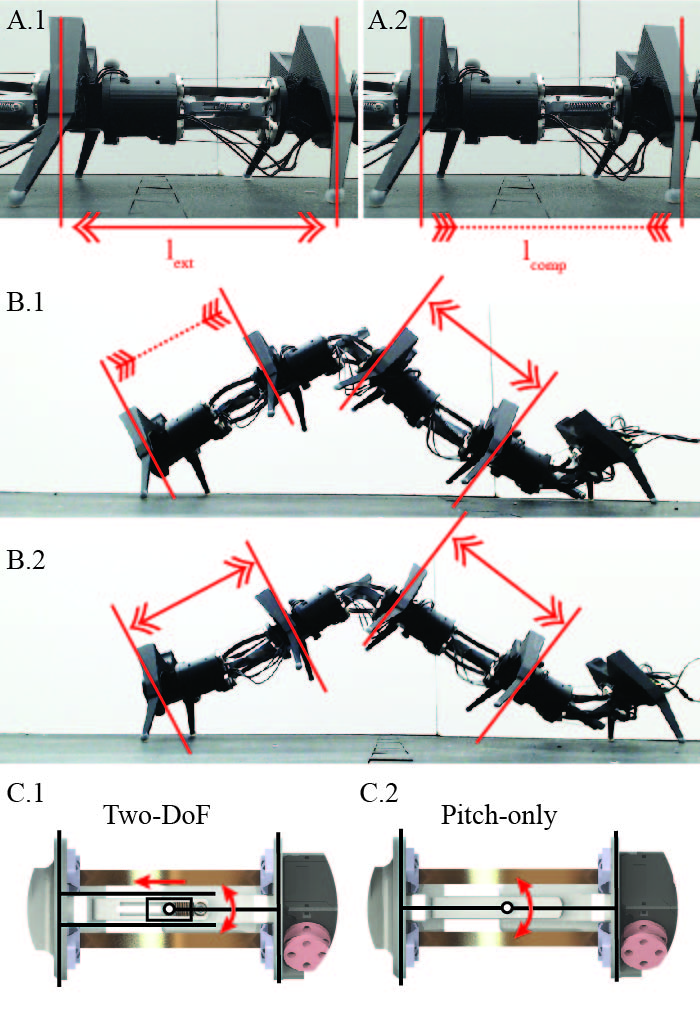}
    \caption{\textbf{Illustration of basic robot capabilities on flat ground.} (A) Snapshots of a joint in (A.1) extension and (A.2) compression states. Compression and extension are labelled with different arrows. (B) Comparison between (B.1) robot using two-DoF joints, capable of propagating a peristaltic wave along the robot body together with the vertical wave, and (B.2) robot using pitch-only joints, capable of propagating a pure vertical wave along the robot body. (C) Comparison between (C.1) the two-DoF joint with both pitch and compression capabilities and (C.2) the pitch-only joint with one rotational DoF in the sagittal plane.}
\end{figure}%

Two stainless steel linear springs, attached on one side to the pin and on the other side to the end of the two-DoF joint slotted component (see Fig.~ 1B), are placed on both sides of each joint. They are required to decouple the two DoFs by providing enough resistance to keep the metal pin in place during a rotation but not during a compression. Finally, we use the spring-steel leaf springs (or spring-steel belts as indicated in \cite{BI2023108436}) to restore the module to the extended state after the contraction, as the cords can only provide stress in the tensile direction rather than in compression. The particular chosen planar configuration for the spring-steel leaf springs allows the pitch joint angle limit to be as large as $[-90^\circ,\ 90^\circ]$.

Each module has a 3D-printed PLA case that houses one Dyamixel 2XL430-W250T (ROBOTIS), which packages 2 independently controlled servo motors. Each servo motor has a pulley (1cm inner diameter) that is spooled with a non-elastic micro cord which has negligible shape memory and deformation response to stretching. The other end of each of the two cords is attached to the legged section of the module. 

Finally, the last section of the module consists of the legged section, made up of two 3D-printed PLA static (non-actuated) legs with hot glue applied at their tips in order to increase the thrust resulting from their passive motion when they interact with the ground. The lower segment of the leg is $15^\circ$ tilted with respect to the upper segment in order to improve stability to the robot.

 \subsection{Power and Communication}

 The robot is powered by a DC power supply with 11.1V and receives control signals transmitted from a PC via U2D2 serial adapter (ROBOTIS). Each servo motor is connected in series with internal wiring running through the modules. Power and communication lines are tied together to create the tether for the robot.

 \subsection{Control and Gait Template}

We control the peristaltic wave and the vertical wave by prescribing the pitch DoF and the peristaltic DoF in each two-DoF joint. 
Specifically, we consider $\alpha(t,i)$ as the $i$th-pitch joint angle at time $t$. Assuming the vertical wave as a serpenoid function~\cite{chong2021frequency}, we have:

\begin{equation}
    \alpha(t,i) = A_{vert} \sin{(2\pi k i/N - \omega t)},
\end{equation}

\noindent where $A_{vert}$, $k$, and $\omega$ denote the amplitude, spatial frequency, and temporal frequency of the vertical waves respectively, while $N$ is the total number of joints. Here, we fix $k=1$ and $\omega=1/2\pi$ throughout each experiment for both waves.

Analogously, we also prescribe the peristaltic wave as a sinusoidal wave. Let $l(t,i)$ be the length of $i$-th peristaltic joint at time $t$. We have:

\begin{equation}
    l(t,i) = l_{ext} - \dl \sin{(2\pi k i/N -\omega t - \varphi)}
\end{equation}
\noindent where $l_{ext}$ is the joint length at the relaxing state, $\dl$ is the magnitude of the compression, and $\varphi$ is the phase offset between the peristalsis and the vertical wave. 

Finally, similarly to \cite{wang2023mechanical}, we calculate the cord length required to achieve the desired angle and compression at each joint at every instant of time by using the following equations:

\begin{equation}
 \begin{cases}
     \mathcal{L}_{upper}  =& [2L_{c}^2+D^2+l^2+2(Dl-L_c^2)\cos(\alpha)\\
     &-2(D-l)L_c\sin(\alpha)]^{1/2}\\
     \mathcal{L}_{lower}  =& [2L_{c}^2+D^2+l^2+2(Dl-L_c^2)\cos(\alpha)\\
    &+2(D-l)L_c\sin(\alpha)]^{1/2}
\end{cases}   
\end{equation}

\noindent where $\mathcal{L}_{upper}$ and $\mathcal{L}_{lower}$ are the length of the upper and lower cables respectively; $D$ and $L_c$ are dimensional constants of the robot (47.5 and 28.5 mm respectively); and $l$ and $\alpha$ are the peristaltic length and the joint angle prescribed by gait equation (see Fig.~\ref{fig:geometric}).

\section{RESULTS}

\begin{figure}
    \centering
    \includegraphics[width=\linewidth]{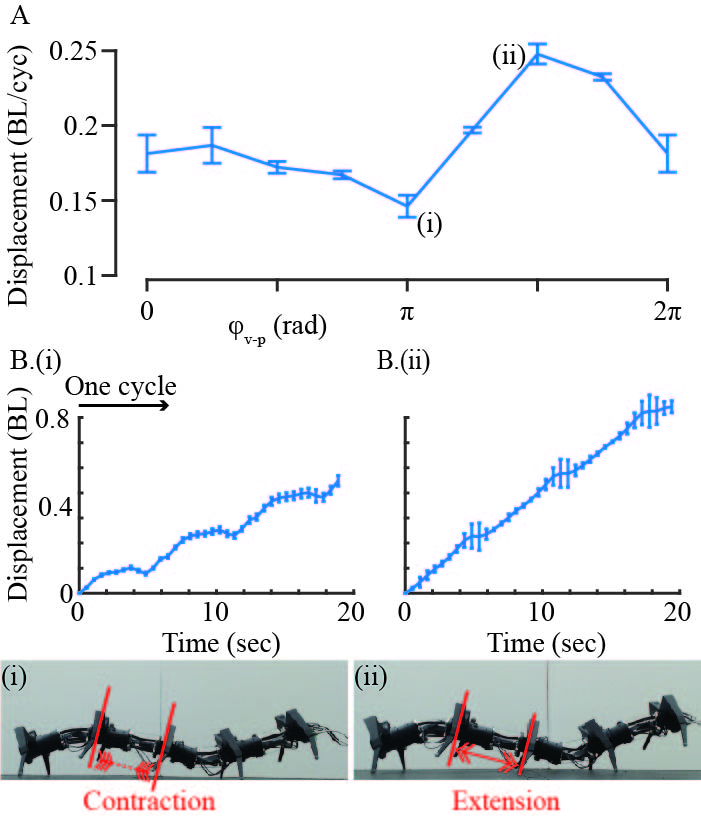}
    \caption{\textbf{Vertical-peristaltic wave synchronization.} (A) Speed (body length traveled per cycle) is plotted as a function of $\varphi$.   Error bars represent standard deviations. (B) (\textit{top}) The displacement as a function of time over three cycles with (i) the inappropriate phase $\varphi = \pi$ and (ii) the appropriate phase $\varphi = 3\pi/2$. (\textit{bottom}) Snapshots of robot performing (i) inappropriately and (ii) appropriately synchronized gait. Detrimental synchronization happens when joint contracts while being lifted and straight. Beneficial synchronization happens when joint extends while being lifted and straight.}
\end{figure}%

\subsection{Flat Ground: Phasing}

In the previous section, we illustrated that the robot is capable of simultaneously generating a peristaltic wave and a vertical wave. We next explored how those two waves can synchronize to produce effective locomotion. Specifically, we quantified the wave synchronization using a single variable $\varphi$, denoting the phase offset between the vertical and the peristaltic wave. Unless otherwise mentioned, we fixed $k=1$, $\omega=1/2\pi$, and $\Delta l=1cm$. Moreover, for simplicity, in this session we fixed the vertical wave amplitude $A_{vert} = 35^\circ$. 

We programmed the robot to implement the prescribed gaits given in Eq.~1 and Eq.~2. In our experiments, we sampled $\varphi$ from $0$ to $2\pi$ with an interval of $\pi/4$. In each trial, we ran the prescribed gait for 3 cycles. At each $\varphi$, we performed 3 trials. 

We tracked the robot real-time position using OptiTrack motion capture system. We illustrate the tracked robot position as a function of time in Fig.~4.B top panel. We used the tracking data to estimate the robot speed (body length travelled per cycle, BL/cyc).

We illustrate the relationship between $\varphi$ and the locomotion speed (mean $\pm$ standard deviation) in Fig.~4A. Variations in $\varphi$ can substantially affect the overall locomotion performance.

In Fig.~4.B bottom panel, we compared snapshots of a robot implementing (i) the  inappropriate ($\varphi = \pi$)  and (ii) the appropriate ($\varphi = 3\pi/2$) vertical-peristaltic synchronization. We note that in (i), the peristaltic-DoF contracts while the joint is straight and lifted off the ground, whereas in (ii) the peristaltic-DoF extends while the joint is straight and lifted off the ground. We posit that these patterns of synchronization can be used by other robots with a peristaltic DoF.

\begin{figure}
    \centering
    \includegraphics[width=\linewidth]{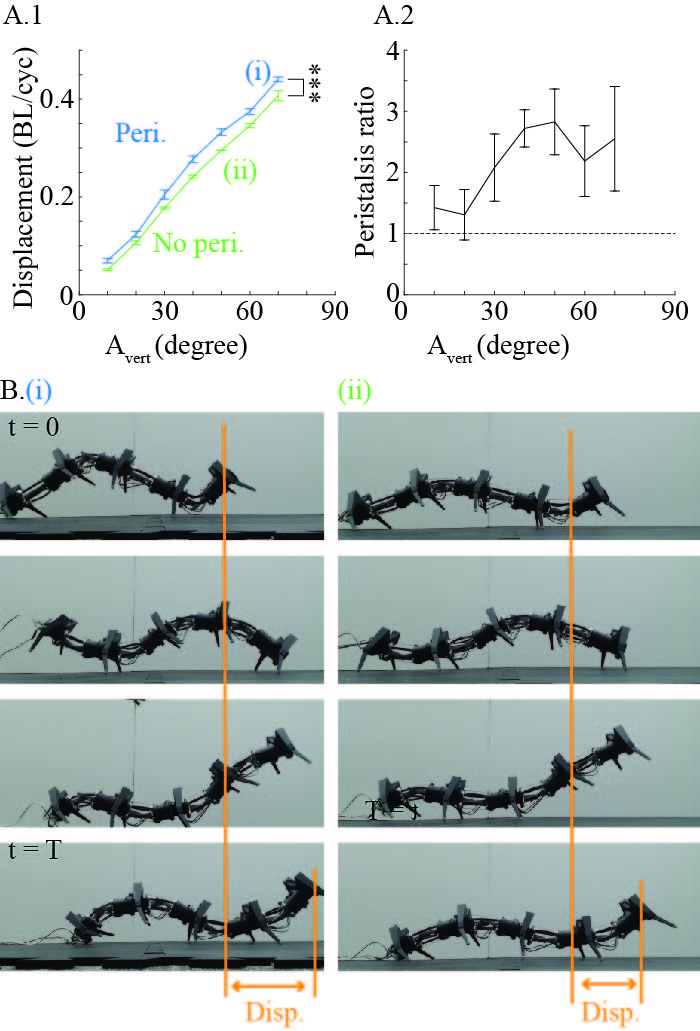}
    \caption{\textbf{Peristaltic wave to improve forward speed on flat ground.} (A.1) Locomotion speed of robot (blue) with peristaltic wave and (green) without peristaltic wave, both as a function of $A_{vert}$. Locomotion speed was measured as the net displacement normalized by the body length of
    the robot over a gait cycle. (A.2) Peristalsis ratio, defined as the speed improvement by peristalsis normalized by the peristalsis wave amplitude, as a function of $A_{vert}$. The peristalsis ratio is significantly greater than 1. (B) Series of time-lapse snapshots over one cycle for (i) $A_{vert}=70^\circ$ and $dl=1cm$, and for (ii) $A_{vert}=70^\circ$ and $\dl=0cm$.}
\end{figure}%

\begin{figure}[h!]
    \centering
    \includegraphics[width=1\linewidth]{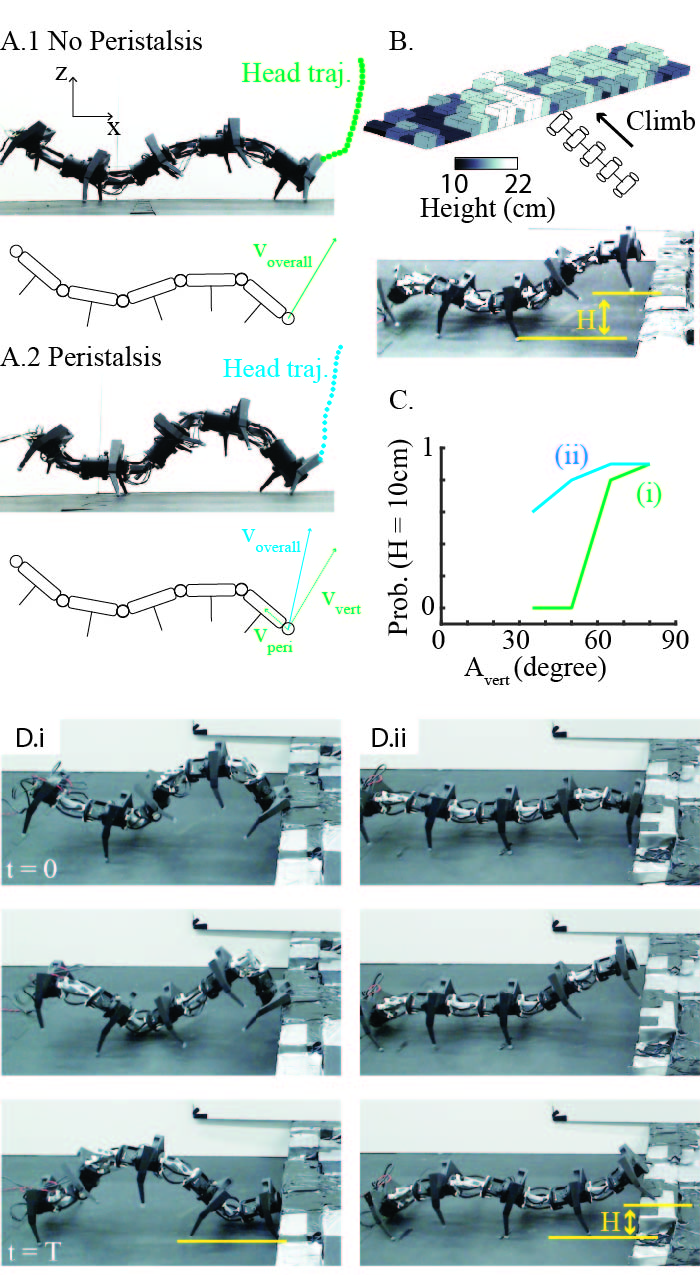}
    \caption{\textbf{Peristalsis wave to improve obstacle-climbing capabilities.} (A.1) (\textit{top}) Snapshot of pitch-only robot version at $A_{vert}=80^\circ$ and $\dl=0cm$. Head movement trajectory is presented in green dots (obtained from experiments). (\textit{bottom}) A diagram illustrating the head velocity components for pitch-only robot locomotion. (A.2) Snapshot of two-DoF robot version at $A_{vert}=80^\circ$ and $\dl=1cm$. Head movement trajectory is presented in blue dots (obtained from experiments). (\textit{bottom}) A diagram illustrating the head velocity components for two-DoF robot locomotion. (B) (\textit{top}) A diagram illustrating the rugose terrain. The terrain has an average height difference of about 0.6 times robot's height. (\textit{bottom}) The edge of the terrain is 10-15 cm from the ground, comparable to the robot's height. (C) Probability of climbing over obstacles for robot (blue, $\dl=1cm$) using a peristaltic wave and (green, $\dl=0cm$) not using a peristaltic wave, both as a function of $A_{vert}$. (D) Snapshot of (i) $A_{vert}=80^\circ$ and $\dl=0cm$ (without peristalsis) failing to climb up the step and (ii) $A_{vert}=35^\circ$ and $\dl=1cm$ (with peristalsis) succeeding to climb the step. Full videos can be found in SI.}
\end{figure}%

\subsection{Flat Ground: Peristalsis ratio}

We next quantified how a peristaltic wave can improve the overall performance on flat ground. We therefore compared the locomotion performance of the two versions of the robot (with pitch-only joints and with two-DoF joints). Note, we keep ($\varphi = 3\pi/2$) during these trials. 

We first tested the robot with two-DoF joints and measured its locomotion speed as a function of $A_{vert}$, the vertical wave amplitude. We sampled a range of vertical wave amplitudes $A_{vert}$ from $10^\circ$ to $70^\circ$ with peristalsis. Snapshots of (i) robot implementing $A_{vert}=70^\circ$ with $\dl=1cm$ are illustrated in Fig.~5.B. We note that locomotion performance increases linearly as $A_{vert}$ increases (see Fig.~5.A.1).

We then sampled the same range of vertical wave amplitudes for the pitch-only joint robot version. Snapshots of (ii) $A_{vert} = 70^\circ$ with $\dl=0cm$ are illustrated in Fig.~5.B. We note that the average locomotion performance is statistically significantly ($p <$ 0.001) slower than for the robot with two-DoF joint (see Fig.~5.A.1). 

To explicitly recognize the contribution from the peristaltic wave, we define a dimensionless number, which we will refer to as the peristalsis ratio: $(D_{p} - D_{np})/\Delta l$, where $D_p$ is the locomotion speed with peristalsis, $D_{np}$ is the locomotion speed without peristalsis, and $\Delta l$ is the amplitude of the peristaltic wave. 

We plot the peristalsis ratio as a function of $A_{vert}$ in Fig.~5.A.2, and we note that this ratio is substantially greater than 1 for all the sampled $A_{vert}$. Given our observation, we hypothesize a much greater speed improvement can result from higher-amplitude peristaltic waves.

\subsection{Rugose Terrains: obstacle-climbing capabilities}
 
We  performed systematic comparative experiments to evaluate the role of peristaltic waves in complex terrain locomotion.

We first built rugose terrains by assembling foam cubes together using glue and duct tape (Fig.~6.B, top panel). The details of the rugose terrain can be found in~\cite{chongScience}. Notably, the edge height of the terrain is 10-15 cm from the ground, similar to the robot height (Fig.~6.B, bottom panel).

In the first experiment, we evaluated the two-DoF joint robot capability to climb over the edge of the rugose terrain, and we compared it with the pitch-only joint robot capability. For both robot versions, we randomly placed the robot close to the edge of the rugose terrain and empirically obtained the probability of the robot successfully climbing over to the rugose terrain within 5 cycles. 10 trials were taken per each combination of $A_{vert}$ and $\dl$. A successful climbing was defined as achieving one pair of legs on top of the rugose terrain. We illustrate examples of (i) unsuccessful and (ii) successful climbing in Fig.~6.D. We calculate the probability of climbing as the number of successful climbing trials normalized by the total trials.

We compare the climbing probability for robots with two-DoF joints and pitch-only joints respectively in Fig.~6.C. The peristaltic wave significantly enhances the probability of successfully climbing the step. Notably, the pitch-only robot version is only able to climb at high amplitudes ($A_{vert}\ge65^\circ$), while when peristalsis is included in the gait the successful climbing probability substantially increases even just at about half such amplitude, ($A_{vert}\ge35^\circ$).

We hypothesize that the peristaltic wave can modulate the head velocity and consequently manipulate its trajectory, which enables the obstacle-climbing capability. We include a diagram in Fig.~6.A. Without peristalsis, the pitch-only joint at the head module will contribute to a velocity to ``raise" the head module (positive velocity projection in z-axis in Fig.~6.A) and climb over the obstacles. However, such velocity has a non-negligible projection in the forward motion (x-axis). As a result, the trajectory of the head movement (Fig.~6.A.1, green trajectory, obtained from experiments) will include a non-negligible ``forward translation", which substantially increases the probability of ``jamming" after collision with obstacles. 

On the other hand, the peristaltic wave will include a velocity component in the backward direction ($V_{peri}$ in Fig.~6.A.2, negative projection in x-axis) due to the module compression. While being small in magnitude, such $V_{peri}$ will substantially reduce the ``forward projection" of the overall head velocity ($V_{overall}$ in Fig.~6.A.2). As a result, the head movement trajectory (Fig.~6.A.2, blue trajectory, obtained from experiments) will have a substantially reduced ``forward translation", which reduces the possibility of ``jamming" after collision with obstacles. 

\begin{figure}
    \centering
    \includegraphics[width=\linewidth]{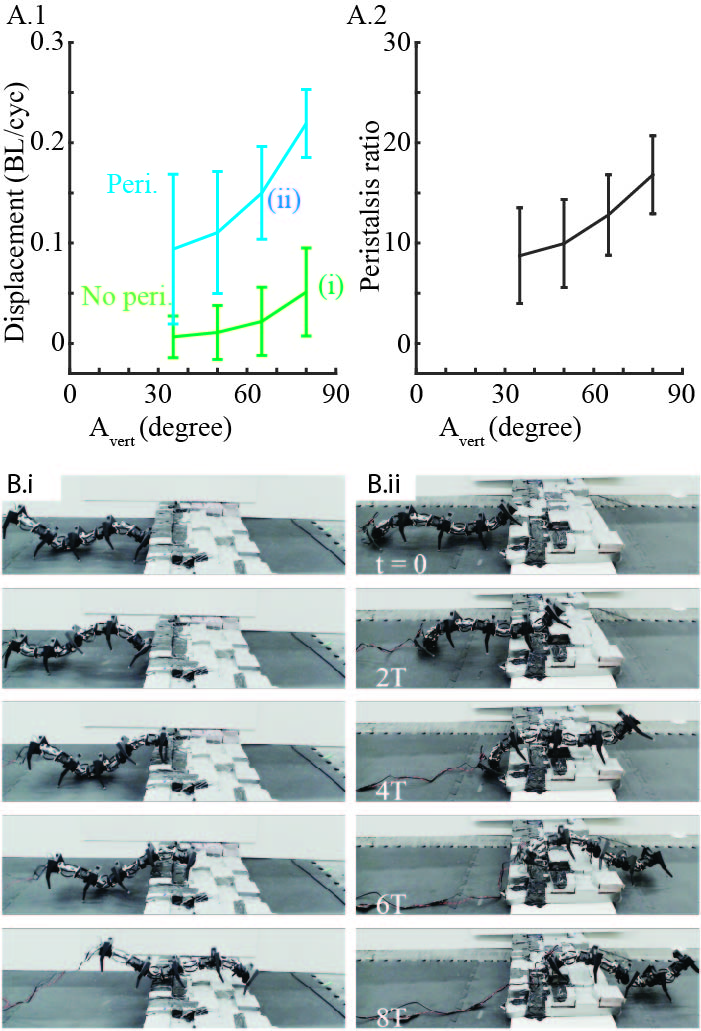}
    \caption{\textbf{Peristaltic wave to improve traversal capabilities on rugose terrains.} (A.1) Locomotion speed on rugose terrain with average height difference about 0.6 times the robot height and with initial step to climb of similar height as the robot, (blue, $\dl=1cm$) using a peristaltic wave against (green, $\dl=0cm$) not using a peristaltic wave, both as a function of $A_{vert}$. (A.2) The corresponding peristalsis ratio, defined as the speed improvement by peristalsis normalized by the peristalsis wave amplitude, as a function of $A_{vert}$. The peristalsis ratio is an order of magnitude greater than the one shown in Fig.~5A.2.  (B) Comparing snapshots of (i) robot at $\dl=0cm$ and $A_{vert}=80^\circ$ traversing rugose terrain and of (ii) robot at $\dl=1cm$ and $A_{vert}=50^\circ$ traversing rugose terrain, after climbing the edge wall.}
\end{figure}%

\subsection{Rugose Terrains: Traversing capabilities}

The second comparative experiment on complex terrains was performed using the same terrain with the same orientation as the first one. In this experiment, the challenges for the robot included (1) climbing the step at the edge of the rugose terrain and (2) consequently traversing the complex terrain. In each trial, we recorded the displacement that the robot can self-transport over the rugose terrain in each cycle (over 10 cycles). We conducted 3 trials at each amplitude, fixing $k=1$, and $\omega=1/2\pi$. We first tested the two-DoF robot with $\dl=1cm$ by sampling $A_{vert}$ from $A_{vert}=35^\circ$ to $A_{vert}=80^\circ$ with an increment of $15^\circ$, then we sampled the same range of vertical wave amplitudes for the pitch-only robot version with $\dl=0cm$.
 
At every value of $A_{vert}$, the two-DoF robot version significantly outperforms the pitch-only robot version (see Fig.~7A.1). We compare the snapshots of the robot traversing the rugose terrain with (i) $A_{vert}=80^\circ$ and $\dl=0cm$ (no peristalsis) and (ii) with $A_{vert}=50^\circ$ and $\dl=1cm$ (with peristalsis) in Fig.~7B. 
We speculate the pitch-only robot suffers from a higher risk of detrimentally interacting with obstacles either in the initial step, or while traversing the bumpy blocks of the terrain, or in both scenarios, which significantly slows the robot down. 
In Fig.~7A.2 we show the peristalsis ratio for different values of $A_{vert}$, and we note that compared to the peristalsis ratio for flat ground locomotion shown in Fig.~5A.2, the peristalsis ratio for complex terrain locomotion is about one order of magnitude larger.

\section*{Conclusion}

In this paper, we identified an alternative actuation mechanism for a multi-legged robot. Specifically, we built a multi-legged cable-driven robot that can locomote effectively on complex terrain by using the combination of a peristaltic wave and a linear progression gait. We showed in comparative experiments that the addition of a peristaltic wave not only enables obstacle-climbing capabilities but also substantially improves the robot's performances when traversing complex environments. Our findings demonstrate how large performance gains can be achieved even just with a seemingly small peristaltic wave amplitude. 

Given our results, we posit that our robot prototype can potentially benefit fields such as search-and-rescue operations, environmental monitoring, and planetary exploration. Future work will study the scalability of this approach, the integration of autonomous sensory systems to further enhance adaptability, and the application of our findings to a broader range of robotic morphologies.

Finally, we note that such performance was not augmented by the leg actuation. However, leg protraction and/or lateral body undulation is not mutually exclusive with our proposed peristaltic wave or vertical wave control schemes. In the future, new explorations are going to be made into the addition of a leg-stepping wave and/or lateral body wave in such a template.

\section*{ACKNOWLEDGMENTS}

We would like to thank the Institute for Robotics and Intelligent Machines for use of the College of Computing basement as a testing space.  
The authors received funding from NSF-Simons Southeast Center for Mathematics and Biology (Simons Foundation SFARI 594594), Army Research Office grant W911NF-11-1-0514, a Dunn Family Professorship, Ground Control Robotics, and a STTR Phase I (2335553) NSF grant. 
We would like to thank Professor Feifei Qian for the fruitful discussion.

\clearpage


\bibliographystyle{unsrt}
\bibliography{bib}

\end{document}